\documentclass[11pt]{article}
\usepackage[margin=1.5in]{geometry}

\usepackage[utf8]{inputenc}
\usepackage[english]{babel}
\usepackage{amsmath}
\usepackage{amsfonts}
\usepackage{color,soul}
\usepackage{graphicx}

\usepackage{subfiles}
\usepackage{subcaption}
 
\usepackage{blindtext}
\usepackage{etoolbox}

\usepackage{main}

\usepackage{booktabs}

\begin{document}

\begin{titlepage}

\centering
{\scshape\LARGE Written Preliminary Examination-II Report \par}
\vfill
{\huge\bfseries {Recent Approaches for Perceptive Legged Locomotion} \par}
\vspace{2cm}
{\scshape\Large {Hersh Sanghvi} \par}
\vspace{1cm}
{\scshape\Large Department of Computer and Information Science\par}
{\scshape\Large University of Pennsylvania\par}
\vfill
{\large May 20, 2022 \par}
\end{titlepage}


\begin{abstract}

As both legged robots and embedded compute have become more capable, researchers have started to focus on field deployment of these robots. Robust autonomy in unstructured environments requires perception of the world around the robot in order to avoid hazards. However, incorporating perception online while maintaining agile motion is more challenging for legged robots than other mobile robots due to the complex planners and controllers required to handle the dynamics of locomotion. This report will compare three recent approaches for perceptive locomotion and discuss the different ways in which vision can be used to enable legged autonomy.

\end{abstract}

\pagebreak

\tableofcontents

\pagebreak

\section{Introduction}
\label{sec:introduction}
Legged robots have unique advantages over other mobile robots. They can traverse stairs and other rough terrain that would stymie wheeled and tracked robots, and they offer greater payload capacities than aerial robots. But even though these robots have the ability to cross rough terrain, they must still carefully plan motions to walk over the terrain without crashing. Designing controllers and planners to take full advantage of these capabilities is still a research challenge because of the hybrid and nonlinear dynamics of these systems. Historically, research has focused on developing blind controllers that do not receive any information about the world around the robot. The most robust of these controllers can handle some rough terrain, but still do not enable robots to cross large obstacles that require significantly altering the robot’s nominal motions like curbs, stairs, large rocks, and more. As legged robots have been increasingly commercialized and deployed for real world tasks, research has shifted towards integrating computer vision and perception into legged locomotion. Methods for perceptive locomotion range in complexity from simply avoiding obstacles to computing joint motions directly from point clouds. In this report, I will provide a brief background on legged locomotion and analyze three works that each are representative of a class of approaches to solving perceptive locomotion. 

The intended audience of this report is practitioners of computer vision who are interested in applications to legged robotics and first year graduate students studying robotics who are interested in control and planning for legged robots.

\subsection{Blind Locomotion}
\subsubsection{Preliminaries}
I will first provide some preliminary terminology and background on the control of legged robots. This section is not meant to be comprehensive, but instead to provide some necessary context for the three papers this report analyzes.

Legged robots move by applying forces to the ground through their feet. These forces, referred to as contact or reaction forces, are transmitted through the legs to accelerate the main body. Although all actuation is done through the legs, tasks and trajectories are often only specified in terms of the main body, or base, of the robot where the center of mass is located. Because these robots are only actuated through the legs, the motion of the base can only be partially controlled. If no legs are in contact with the ground, then the position and linear velocity of the base cannot be controlled at all. In addition to this underactuation, there are many other factors that legged robots must account for to walk successfully. The body could fall over due to an unstable control action. Any one of the many links of the robot could collide with the environment, which could destabilize the robot. If a foot applies too much force, it loses sticking contact with the ground and therefore can no longer effectively support the body. All of these factors must be considered when designing a control strategy.

Most legged robots of interest have multiple legs; two of the dominant design patterns in modern legged robots are quadrupeds (robots with 4 legs) and bipeds (robots with 2 legs). The coordinated patterns by which the legs move are called \textit{gaits}, which are predefined sequences during which certain legs are in contact with the ground, referred to as \textit{stance} or \textit{support} phase, while others will be moving through the air, referred to as the \textit{swing} phase. These gaits are often modeled after the way animals move. By changing the phase durations and phase offsets between legs, different biological gaits, such as crawling, trotting, and galloping, can be created.

Historically, controllers for legged robots have aimed to generate contact forces for stable walking behaviors while following a certain gait pattern. For quadrupedal robots, as long as the footprint of the center of mass on the ground is within the polygon defined by the feet in the stance phase, the robot will not fall over \cite{mcghee_stability_1968}. This means that for a crawling gait, where only one leg is ever in swing phase, a quadruped will always be stable as long as the stance feet locations are chosen properly. However, such statically stable motions are typically very slow; most dynamic motions for bipeds and quadrupeds are naturally unstable.

\subsubsection{Dynamic Models}
The continuous portion of the rigid-body dynamics of legged robots can be derived from Newtonian or Lagrangian dynamics:
\begin{align} 
    \label{eqn:multibody}
    M(q)\ddot{q} + \dot{q}^\top C(q) q &= S^\top \tau + \tau_g(q) + \sum_{i=0}^{k} J_{C_i}^\top f_i
\end{align}

Where $q$ are the generalized coordinates of the robot (the center of mass state and joint states), $M(q)$ is the inertia matrix, $C(q)$ is the Coriolis force, $S$ is a matrix that selects the coordinates that represent actuated joints from $q$, $\tau$ is a vector of torques at those joints,$\tau_g$ is the gravitational force at each link, $J_{C_i}$ is the Jacobian matrix of a point $C_i$ that is in contact with the ground, and $f_i$ is the force exerted by the ground onto that point.

Due to the complexity and nonlinearity of the multibody dynamics (\ref{eqn:multibody}), many simpler models have been developed for planning and control. One common approximation for the stance dynamics is the Single Rigid Body (SRB) approximation, which, following the derivation from \cite{winkler_optimization_2018}, ignores the actuated subsystem (ie. the motion of all actuated joints) by assuming that the torque limits are large enough to make controlling them trivial. Furthermore, SRB assumes that the motion of the actuated joints do not impact the momentum of the robot (ie. they have very low mass) and that the inertia of the robot does not depend on its joint configuration. This effectively reduces the model to a set of forces (from contacts) projected onto the base of the robot. By separating the linear and angular terms, this gives us the following equations for the dynamics of the base:
\begin{align}
    m\ddot{p} &= mg + \sum_{i = 1}^{k} f_i \\
    I(\theta)\dot{\omega} + \omega \times I(\theta) \omega &= \sum_{i = 1}^{k} f_i \times (r_i - p) 
\end{align}

The first equation is the translational Newtonian dynamics, where the sum of the total forces (from gravity $g$ and $k$ contact forces $f_i$) is equal to mass times acceleration. The second equation is the rotational dynamics, where $I(\theta)$ is the moment of inertia, $\theta, \omega$ are the orientation and angular velocity, and $r_i$ is the location of the $i$th contact point.

This dynamic model can be used in optimization-based strategies like Model Predictive Control (MPC), and further approximations can be made to speed up the optimization problem and render it convex. For example, the cross product between forces and footstep locations in (3) would create a nonlinear constraint. This problem is further compounded if searching for footstep locations on nonconvex terrain. Thus, footstep locations are not treated as decision variables and are instead pre-specified based on some other criterion. 

Once the reaction forces have been computed from the SRB model, they can be fed into another lower-level controller that computes the necessary motor torques to be applied at each joint to achieve the desired reaction force. Many legged robot control stacks, including the ones analyzed in this report, have this hierarchical structure where an outer loop controller computes reduced-order setpoints using simplified models at a lower rate, and then inner-loop controllers calculate high-bandwidth motor torques using more detailed knowledge of the robot’s geometry and state. 

Further approximations have also been made on top of the SRB. One common template \cite{full_templates_1999} for legged locomotion is the spring-loaded inverted pendulum (SLIP) model \cite{blickhan_spring-mass_1989}, which models the robot motion as that of a point mass attached to a massless sprung leg. This model was developed by observing the running motions of animals, and has been used to define target invariant surfaces for higher-dimensional legged robots \cite{poulakakis_spring_2009}. Another reduced-order model is the 3D Linear Inverted Pendulum Mode (LIPM) \cite{kajita_lipm_2001}, which casts the walking motion of the robot as a point mass with a constant height with no angular velocity supported on a telescoping, massless leg. The control strategies developed for these simple SLIP and LIPM models are still used on high DoF robots, particularly for selecting footstep locations \cite{kim_highly_2019,gibson_terrain_2022}. 

\subsubsection{Learning-Based Methods}
Although these models are useful for planning and control, they do not capture the full capabilities of the robot and thus restrict the potential agility of the robot. Recently, using machine learning (ML) for locomotion has become a viable approach. ML has long been used in domains like computer vision (CV) and natural language processing where closed-form functions from the inputs to outputs are almost impossible to obtain. Instead, ML enables learning a parametric model from data. In the context of legged locomotion, ML could be used to learn a dynamics model or to directly learn a controller model. ML is appealing for legged robots as it could eliminate the dependence on reduced order models like SLIP and SRB and learn more expressive models and controllers with data from the real robot or a high-fidelity physics simulator. In addition, with modern compute, training and inference for learned models has grown faster, allowing ML to be used in medium to high bandwidth applications like control. ML, and deep learning in particular, have been applied to speed up optimal control for legged robots by learning value functions \cite{deits_lvis_2019}, learning good initializations for optimization problems \cite{melon_reliable_2020}, and learning to classify feasible contact state transitions \cite{lin_efficient_2019}. 

Reinforcement learning (RL), a subfield of ML, has received a great deal of attention within the legged locomotion field. In the RL paradigm, at each timestep, a policy observes the state of some environment, which could be dynamical states of a robot, locations of obstacles or enemies, images from a video game, and more. Then, it has to choose an action based on that state. After selecting an action, the policy will receive a reward and observe the next environment state. Informally, the reward gives the policy information about its performance; a high reward indicates desirable actions and states. 

RL can be model-based or model-free; model-based RL aims to learn a model that can then be used to plan, while model-free RL aims to directly learn the policy. The goal of RL is to learn a policy or model which selects actions that maximize the sum of expected future rewards from the current state. RL does not require having expert data to learn from, which is desirable for locomotion tasks in particular due to the difficulty of designing good controllers for arbitrary terrains. The designer only has to specify an action space, observation space, and the reward function. In practice, the selection and tuning of these has a huge impact on the performance of the trained policy, especially for locomotion. This informal explanation discards many of the intricacies and practical challenges of RL; for a more in-depth tutorial, refer to the seminal book \cite{sutton_rl_1998} or \cite{li_deep_2018} for more modern approaches.

Many modern RL methods use deep neural networks to parameterize the policy; this is referred to as Deep Reinforcement Learning. Typically, deep RL methods cannot be used to train policies using real robots because they require huge amounts of training samples to learn useful policies, and collecting this data on a real robot would take a prohibitively long time. Furthermore, during the training phase, RL policies employ exploration, where they select random actions to find potentially more optimal policies. This exploration usually results in collisions with the environment, falling over, and pushing joints past their limits. As a result, training an RL policy on a real robot is generally avoided. As an alternative, fast and accurate physics simulators are employed to train policies fully in simulation. However, RL methods trained in simulation generally suffer from the reality gap (also called the sim2real problem). When the policy (trained in simulation) is deployed on a real robot it fails to produce optimal, or even stable, behaviors due to  mismatches between the simulation dynamics and real robot dynamics. We will examine some common solutions \cite{peng_sim_2018} in more detail in the summaries of the papers.

Despite the sim2real problem, RL-based methods have shown strong performance on many locomotion tasks on real robots. Recent work demonstrated robust RL-based policies for walking on flat ground and stairs with the Cassie bipedal robot \cite{xie_feedback_2018,siekmann_blind_2021}. Deep RL-based controllers have been used on the ANYmal-C quadrupedal robot to enable robust walking and trotting on a large variety of rugged outdoor and indoor terrains \cite{lee_learning_2020}. Adopting a slightly different paradigm, controllers that adapt to estimated changes in the environment using latent state estimation or meta-learning have also been trained in simulation using deep RL and deployed on the Unitree A1 \cite{kumar_rma_2021} and Minitaur quadrupeds \cite{song_rapidly_2020}. Many of these methods take a long time to train due to their sample inefficiency, but solutions to this have also been developed. Using massively parallel simulators, RL algorithms can gather millions of training samples in as little as 20 minutes to train a locomotion policy on a variety of legged robots \cite{rudin_learning_2021}. Sample-efficient deep RL training methods have also been used to train the Minitaur robot to walk in limited indoors environments without relying on simulation at all \cite{haarnoja_learning_2019}. Most of the approaches mentioned thus far have been model-free, but deep RL has also been applied to learn dynamics models on real robots in as little as 5 minutes \cite{yang_data_2019}.

\subsection{Perceptive Locomotion}
Although blind locomotion strategies have achieved high performance on various locomotion tasks, they are limited in terms of their robustness. In order to avoid undesired collisions between the legs or body and the environment, some information about the external world is needed in the form of a map, depth images, or point clouds. Seamlessly integrating this information into traditional legged robot control stacks is still a challenge. Planners and controllers using exteroceptive information must be able to account for the complex dynamics of the robot and nontrivial interactions with the surrounding terrain while still running quickly enough to be used online. It is also not obvious a priori what kind of mapping from perception to action should be used, nor how much of the robot’s action selection should be based on perception. This has led to a large variety of approaches to incorporating perceptive information. These range from feeding terrain heightmaps into a learned recurrent model to generate long-horizon footstep plans for a one-legged hopper \cite{hsanghvi_fast_2021}, to using a traditional path planning algorithm along with an SRB to plan global CoM trajectories \cite{norby_fast_2020}, to using onboard RGB cameras to classify different terrain textures and planning to avoid softer terrains \cite{wellhausen_where_2019}. 

In this report, I will summarize and analyze three papers that each shows a different way of integrating perception into a legged robot control stack. Each paper is broadly representative of a common type of approach present in the modern literature. In section \ref{sec:comparison}, I will compare these papers in terms of a few key metrics: how they create an information bottleneck from perception to control, how they deal with perception uncertainty, what kinds of terrain they can handle and how autonomous they are, and the engineering effort involved in implementing each method. In Section \ref{sec:futurework}, I will explain how these methods could be expanded upon in the future. Lastly, I conclude in Section \ref{sec:conclusion} by summarizing some takeaways that appear across all three papers.

\section{Vision Aided Dynamic Exploration of Unstructured Terrain with a Small-Scale Quadruped Robot \protect\cite{kim_vision_2020}}
\label{sec:visionaided}
This work \cite{kim_vision_2020} develops a high-level perceptive planner that extracts gradient features from the perceived heightmap to plan footsteps and avoid obstacles. This perception module is layered on top of a blind optimization-based motion controller. The planner and controller are tested on the small-scale MiniCheetah quadruped. Similar approaches in the literature include \cite{kagami_vision_2003,kalakrishnan_learning_2009,wahrmann_fast_2016,magana_fast_2019,villarreal_mpc-based_2020}, all of which use either learned or hand-specified features from the perceived terrain to plan footsteps and avoid obstacles with a low-level controller. This section will go over the blind motion controller, the perception system and its design considerations, and finally some analysis on the strengths and weaknesses of this approach.

\subsection{Motion Controller}
The low-level controller used in this work was first presented in \cite{kim_highly_2019}. Since the details of this controller implementation are mostly out of scope for this report, I refer the reader to \cite{kim_highly_2019} for the mathematical details. In this section, I will give a high-level overview of the controller scheme.
The motion controller consists of a Model Predictive Control (MPC) planner that outputs desired ground reaction forces for the current stance phase composed with a Whole Body Impulse Controller (WBIC) that calculates the necessary joint torques to achieve those ground reaction forces. The WBIC uses a more accurate dynamic model, but is limited to running for only one timestep forward due to bandwidth constraints. Meanwhile, the MPC uses the SRB model, which is less dynamically accurate, but consequently can be run faster and find reaction forces for an entire gait cycle, allowing some planning to be done.

\subsubsection{MPC and Footstep Selection}
The MPC takes as input a desired center of mass (CoM) trajectory and gait schedule, and outputs the forces the robot needs to apply at its feet in order to follow that trajectory during the upcoming gait cycle. Rather than just outputting a dynamically feasible center of mass trajectory, reaction forces are used as the intermediate signals between the MPC and the lower level WBIC because they are more feasible to follow than CoM trajectories for gaits that have long unactuated phases, such as galloping where all four feet are in swing phase for a significant portion of the gait cycle.

The MPC is formulated as a quadratic program (QP), with dynamics, initial state, and friction cone constraints. However, the SRB dynamics are not linear without additional assumptions. In order to further simplify the problem and make the dynamics linear, the authors make a small angle approximation on the pitch and roll of the body, which cancels out the cross product terms. Furthermore, they assume that the states are close to the commanded trajectory, thus simply using the commanded yaw rotation $R_z(\psi_k)$ as the orientation in the dynamics. The pitch and roll velocities of the CoM are assumed to be small in order to eliminate the cross product term from the angular velocity calculation. Therefore, the dynamics become linear, and the MPC is framed as solving the following QP:

\begin{align*}
    \text{minimize}_{x,f} & \quad \sum_{k = 0}^{T} \|x(k+1) - x^{\text{ref}}(k+1)\|_{Q} + \|f(k)\|_{R} \\
    \text{subject to} & \quad \text{linearized SRB dynamics (2) and (3)} \\
    & |f_x(k)| \leq \mu f_z \quad \forall k \\
    & |f_y(k)| \leq \mu f_z \quad \forall k \\
\end{align*}

Where $x$ is a vector of the CoM position, rotation, velocity, and angular velocity, and $f$ are the reaction forces. $x^{\text{ref}}$ is the commanded trajectory, and $Q,R$ are state and input cost matrices in the LQR sense. The last two constraints are linear friction cone constraints to ensure the feet in contact do not slip. 

An important component of keeping the MPC formulation linear is having a fixed gait schedule and known footstep locations. Before the MPC controller is called, a periodic gait scheduler determines which feet are in swing and stance during the planning horizon, and reports the contact states to the MPC. The upcoming stance foot locations $r_i$, which remain fixed during the stance phase, are calculated based on the commonly-used “Raibert" heuristic, from Raibert’s work on the control of one-legged hopping robots using the SLIP template (covered in Section \ref{sec:introduction}). The footstep location is calculated using P control based on the difference between current and desired linear velocity of the base:
\begin{align*}
    r_i &= \frac{t_{\text{stance}}}{2}v + k(v - v^{\text{cmd})}
\end{align*}

The first term is the Raibert heuristic for a nominal footstep location that will not alter the velocity of the base, and the second term is a proportional term that modifies the step location based on the difference from the setpoint.

\subsubsection{WBIC}
The SRB MPC has no notion of the geometry of the robot or how the robot should achieve the commanded reaction forces. The WBIC is used to find the joint position, velocity, acceleration, and torque commands to achieve those reaction forces and other specified tasks like swing foot control. The joint position, velocity, and acceleration commands are computed iteratively for each “task”, following a strict priority order. The Jacobian of each task (ie. how task space quantities change with joint angles) is projected into the null space of the tasks with higher priorities. Essentially, this describes how each task can be executed without interfering in the execution of tasks with higher priorities. The tasks in this work are to follow the commanded body orientation, body position, and foot position, in that order. The exact computation of this procedure is out of the scope of this report, but it can be thought of as outputting joint position and velocity targets to a PD controller for each joint, and joint acceleration commands are fed into another QP that uses the locally linearized full-order dynamics (\ref{eqn:multibody}) to compute the torque command for a single timestep:
\begin{align*}
    \label{eqn:multiopt}
    \text{minimize}_{\delta} \quad & \delta^{\top}_{f_r} Q_1 \delta_{f_r} + \delta^{\top}_{f}Q_2 \delta_{f} \\
    \text{subject to} & \quad  S_f(A\ddot{q} + b + g) = S_fJ_c^\top f_r \\
    & \ddot{q} = \ddot{q}_{\text{cmd}} + \delta_f \\
    & f_r = f_r^{\text{MPC}} + \delta_{f_r} \\
\end{align*}

Where $A, b, g$ are linearized versions of the terms in (\ref{eqn:multibody}), and $S_f$ selects the base states from the generalized coordinates. $J_c$ is the contact Jacobian. The latter two constraints frame this problem as computing relaxations to the commanded floating base accelerations and reaction forces. This optimization problem computes deltas to the MPC reaction forces to make them consistent with the desired task accelerations and the full-body dynamics model. The full-body dynamics models the effect of joint masses, Coriolis force, and reaction forces on the acceleration of all joints. From the optimized reaction forces, the feedforward joint torques are computed from the full dynamics and added to the output of the PD controller, and are sent to the motors.

This blind motion controller system is able to achieve indoor running on a treadmill at 3.7 m/s on a treadmill, which is very agile relative to the size of the Mini-Cheetah platform. Outdoors, the controller enables Mini-Cheetah to run at 1 m/s on a variety of surfaces including grass and gravel. Based on an analysis of the joint torques and velocities, the controller is able to utilize the hardware to nearly its maximum capacity.

\subsection{Perception System}
\begin{figure}
    \centering
    \includegraphics[width=0.5\textwidth]{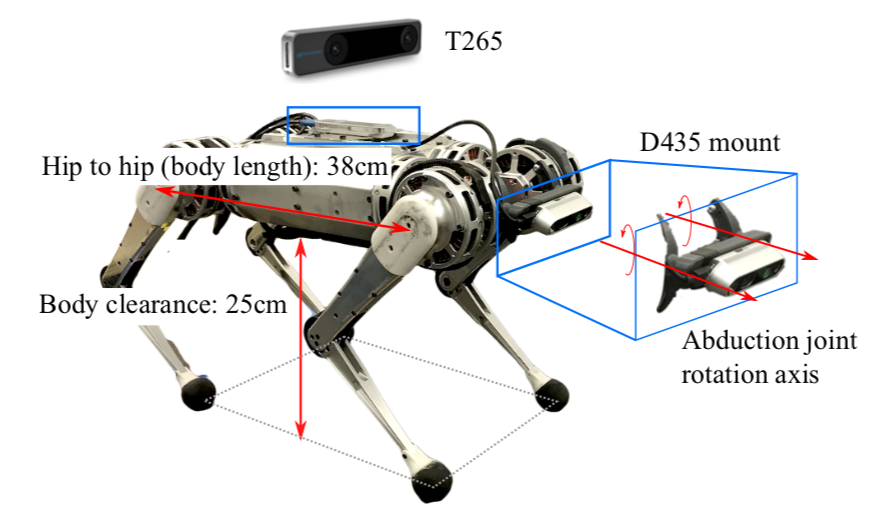}
    \caption{The MiniCheetah quadrupedal robot with depth cameras attached.}
    \label{fig:minicheetah}
\end{figure}

\begin{figure}
    \centering
    \includegraphics[width=0.5\textwidth]{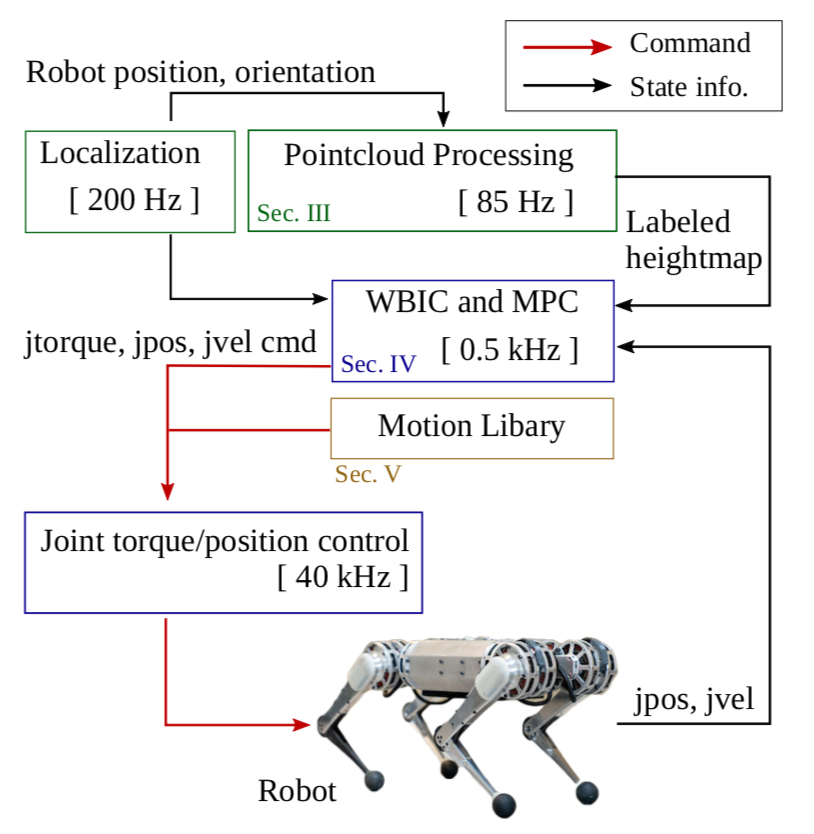}
    \caption{The system architecture from \protect\cite{kim_vision_2020}}
    \label{fig:minicheetah_sys}
\end{figure}

The design of the perception system in this work is highly guided by the packaging size constraints of the Mini-Cheetah quadrupedal robot, shown in Figure \ref{fig:minicheetah}. Due to its small size, more accurate, yet bulky, sensors like LIDARs cannot be used on the robot. In addition, there is limited space for an onboard computer. Therefore, the designers elect to use two small Intel Realsense cameras as perceptive sensors. The D435 is used for the construction of a map, while the T265 is used for its SLAM and localization capabilities. Their overall system architecture is shown in Figure \ref{fig:minicheetah_sys}.

Notably, perception is not done fully onboard. While the localization through the T265 is done onboard the robot, the point cloud processing occurs on an external desktop. The authors state that this is a packaging restriction rather than a computational one; and their future work \cite{dudzik_robust_2020} indicates that this setup does not actually require an external desktop.

The camera-frame point clouds that are produced by the D435 are transformed into the world-frame (using the estimated pose of the robot from the T265) so they can be used for footstep planning and obstacle avoidance. The point clouds are used to construct a 2.5 dimensional heightmap, which is an image where each pixel corresponds to an (x,y) location in the real world, and the value of the pixel is the z coordinate of the terrain at that location. The heightmap is then filtered to remove noise, using an erosion operation which removes outliers, followed by a dilation, which fills in regions which might be sparsely sensed in the original point cloud. This filtering process alleviates some of noise from the stereo camera.

Once the filtered heightmap in the local vicinity of the robot is computed, it is used to classify the terrain into 4 buckets: steppable, unsteppable, jumpable, and impassable. Steppable terrain, as the name suggests, is flat terrain that can be crossed without problems. Unsteppable terrain corresponds to regions where it would be dangerous for the robot to place its feet. Jumpable terrain is terrain that can be reached by engaging a hardcoded jump motion, and impassable terrain is one that cannot be crossed at all. The terrain is classified using features extracted by convolving the heightmap with vertical and horizontal Sobel kernels:
\begin{align*}
    S_x &= \begin{bmatrix}
    -1 & 0 & 1 \\
    -2 & 0 & 2 \\
    -1 & 0 & 1 \\
    \end{bmatrix} \\
    S_y &= \begin{bmatrix}
    -1 & -2 & -1 \\
    0 & 0 & 0 \\
    1 & 2 & 1 \\
    \end{bmatrix}
\end{align*}
which essentially computes vertical and horizontal gradients of the heightmap. Each region of the image is classified as unsteppable, jumpable, or impassable depending on the magnitude of the gradient of that region. Flatter regions of the terrain are classified as steppable.

This classified terrain is used for footstep planning. Footstep locations are first calculated in the same way as in the low-level controller in the prior section. However, if the nominal footstep lands in an unsteppable or impassable location, the heightmap is searched in a spiral pattern for the nearest safe step location. Once a safe footstep has been found, the heightmap is also queried to find the z height of the new footstep, and the expected landing height and swing foot trajectory are adjusted accordingly in the controller.

In order to get obstacle-avoidance behavior, when computing velocity commands, a radial basis function potential field is computed around impassable regions. The gradient of this field is added to any user-commanded velocity, autonomously steering the robot away from obstacles while still closely following a path computed from a global map or commanded by the user.

\subsection{Experimental Evaluations}

\begin{figure}
    \centering
    \includegraphics[width=0.5\textwidth]{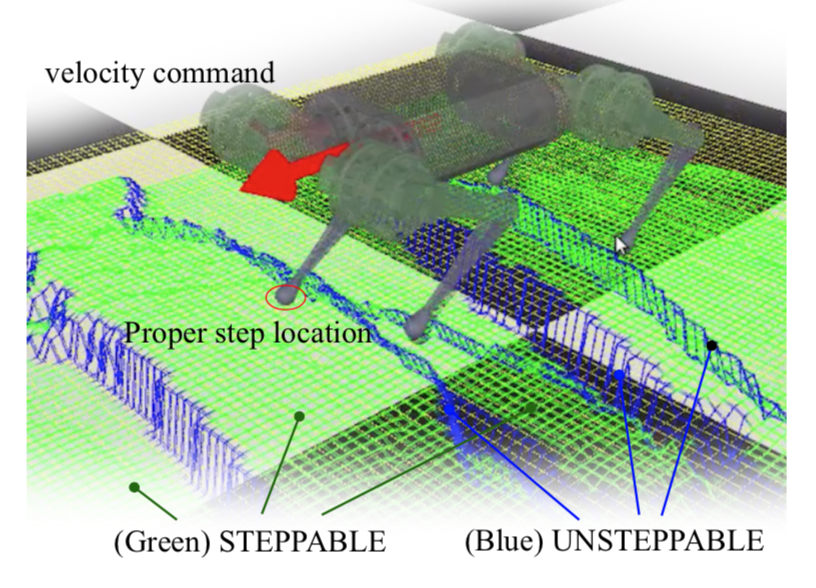}
    \caption{Footstep adjustment based on perception from \protect\cite{kim_vision_2020}}
    \label{fig:minicheeta_stepclass}
\end{figure}

With the addition of perceptual input, the three-stage control pipeline enables the MiniCheetah to avoid both static and moving obstacles without explicit operator intervention, and to cross rough terrain in the form of some stacked planks The footstep planning system avoids stepping on the edges of the planks which are stacked up to 10 cm high, which is half the robot’s ground clearance. An example of their footstep planner in action on the planks is shown in Figure \ref{fig:minicheeta_stepclass}. The authors' future work also shows this vision-aided pipeline enabling stair climbing in simulation \cite{dudzik_robust_2020}. The perception system allows the robot to trot at 0.33 m/s over these planks, which is unfortunately slow compared to the capabilities of the low-level controller on flat terrain, but comparable to the other works in this report.

\subsection{Conclusion and Analysis}

The low level motion controller provides a strong and robust baseline on top of which to add perceptive input. This design pattern will appear in the upcoming papers as well. The perception system will not be fully reliable in all cases, so having a robust underlying controller is critical to the performance of the overall system. In this case, the strength of the motion controller allows a relatively simple perception system that is only used for footstep planning and broad obstacle avoidance. However, the MPC and WBIC implementations require accurate state estimation and computation of the local full-body dynamics, which are nontrivial. The MPC QP, Kalman filter-based state estimation, Jacobian null-space task control, dynamics computation, and WBIC QP must all run online on an embedded computer at a high rate. In addition, approximations such as the small angle one made for the linearized SRB dynamics break down when the robot is on an incline or climbing stairs. 

This work is similar in flavor to a number of other locomotion works which find a cost map, either through machine learning or some simple heuristics, on the terrain in order to modify nominal footstep locations. One shortcoming of this work relative to the others is that the kinematic workspace of the robot is not considered while planning footsteps. Nominally, if there is a ditch in the ground, their gradient computation through Sobel filtering would show its edges as impassable, but not the ditch itself. This is not a critical flaw of the method, as it would be simple to add a heuristic to the heightmap that does not allow stepping on regions below a certain depth. None of these are insurmountable obstacles, but each requires careful consideration when designing the system.

A major advantage of this method is that the perception system is relatively simple to implement, especially since localization is performed using the T265 camera, an off-the-shelf component. The only robot-specific tuning factors to implement in the perception system are the gradient thresholds for the different terrain classifications and the gait schedule. Therefore, if one already has a low-level controller implemented for the legged robot, this type of approach is an easy way to add perception while retaining its capabilities. Additionally, the terrain classification system allows using a library of offline-optimized motions, which can be a powerful paradigm for agile locomotion. 

On the other hand, the relatively simple perception integration is limited in how much it can improve on the low level controller. For example, although perception can be used to query the footstep $z$-heights, it cannot alter the gait at all, meaning that no matter how rough the terrain, the robot is restricted to a predefined gait schedule. Furthermore, although the filtering and mapping systems may work well indoors, such hand-engineered systems can break down in more rugged outdoors environments due to lighting variations. 

Overall, this work presents a compelling design pattern by integrating a stable and fast low level controller with a relatively simple perception system. However, the requirements for a global mapping scheme and relatively simple perception to action mapping limit how much the perception system can improve the capabilities of the low level controller. More analysis and comparisons with the other methods are presented in Section \ref{sec:comparison}.

\section{Visual-Locomotion: Learning to Walk on Complex Terrains with Vision \protect\cite{yu_visual_2022}}
\label{sec:visuallocomotion}
This work by \cite{yu_visual_2022} presents a similar approach to the previous one, with perception providing high-level commands to a model-based low-level controller. But instead of using terrain heuristics, this paper uses RL to learn the mapping from depth images and base states to high level commands. This is similar to \cite{villarreal_mpc-based_2020}, \cite{gangapurwala_rloc_2020}, and \cite{melon_recedinghorizon_2021} which use a learned high-level policy using perception in concert with a model-based low level controller.

    \subsection{System Architecture}
    \begin{figure}
        \centering
        \includegraphics[width=0.5\textwidth]{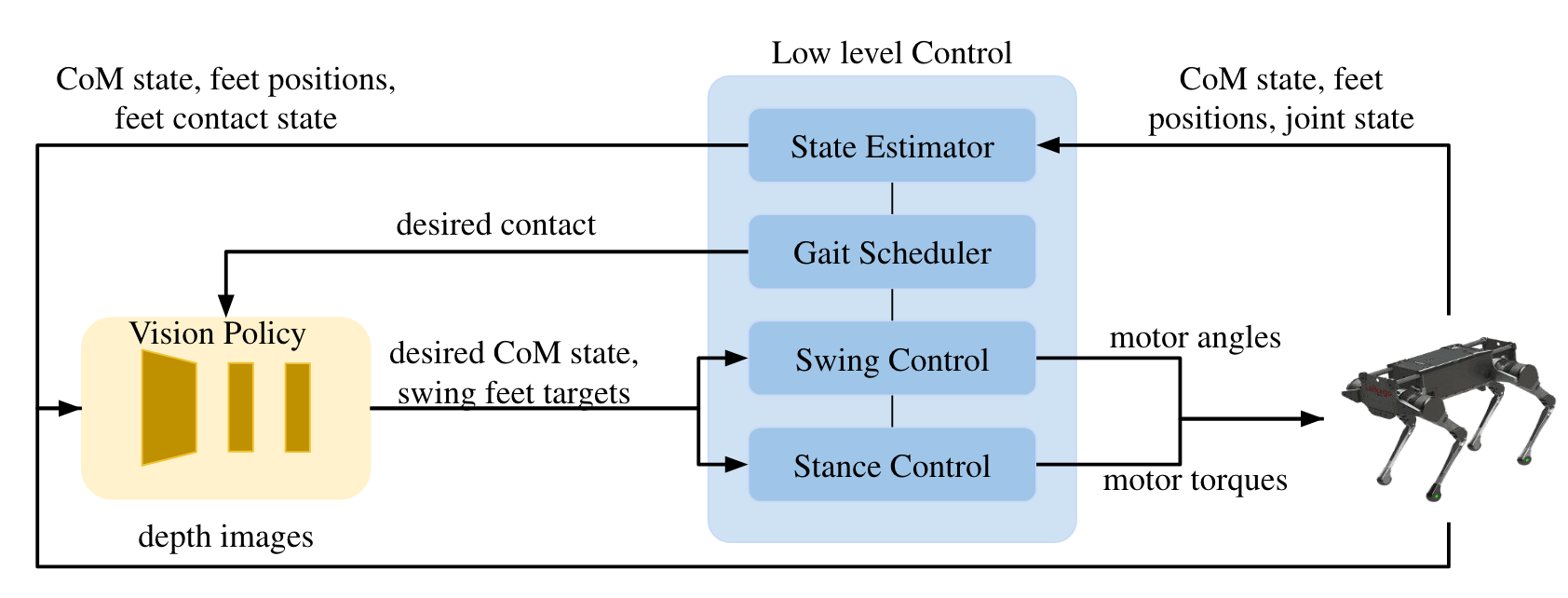}
        \caption{The system architecture of \protect\cite{yu_visual_2022}}
        \label{fig:visualloco_sys}
    \end{figure}
        \begin{figure}
        \centering
        \includegraphics[width=0.5\textwidth]{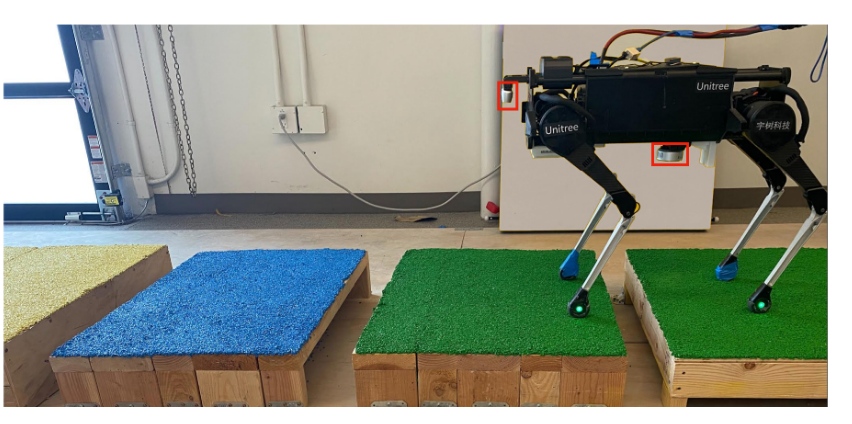}
        \caption{The Unitree Laikago robot used in \protect\cite{yu_visual_2022}}
        \label{fig:visualloco_robot}
    \end{figure}

The hierarchical system architecture presented in this paper is shown in Figure \ref{fig:visualloco_sys}. The visual inputs, along with the proprioceptive state of the robot and the desired contact state, are jointly processed by a learned policy that outputs desired positions for the CoM and upcoming stance foot, and a velocity for the CoM. The output of the learned policy is fed as a target to an SRBD-based MPC that outputs desired stance foot forces, which are in turn converted into desired motor torques. The swing feet are controlled to follow a pre-specified trajectory using inverse kinematics. The vision policy runs at 20Hz, while the MPC controller runs at 250 Hz. The approach is tested on a Unitree Laikago robot (Figure \ref{fig:visualloco_robot}), a quadrupedal robot weighing 24 kg with 12 actuated joints. 

Two depth cameras provide visual inputs. One, an Intel D435 stereo camera, is attached to the front of the robots and provides a wide-angle view of the terrain in front of the robot. The second, an Intel L515 Lidar-based camera, is attached to the robot’s underside and provides a high-resolution view of the terrain directly underneath the robot.

\subsection{Low-Level Controller and Gait Scheduler}
The low-level controller used in this work is similar to the one used in the previous paper. The controller uses MPC based on the SRB approximation to compute the desired contact forces over the prediction horizon while also enforcing maximum force and friction cone constraints. However, instead of doing a further optimization to compute motor torque commands, they multiply the reaction forces by the transpose of the contact Jacobian to get the torques.
    
For swing foot control, there is a pre-specified swing foot trajectory that each foot tracks using inverse kinematics, where the desired end-effector position is converted to desired joint angles and those desired angles are fed to a PD controller, which calculates the motor torques necessary to hit the target joint angles.

\begin{figure}
    \centering
    \includegraphics[width=0.75\textwidth]{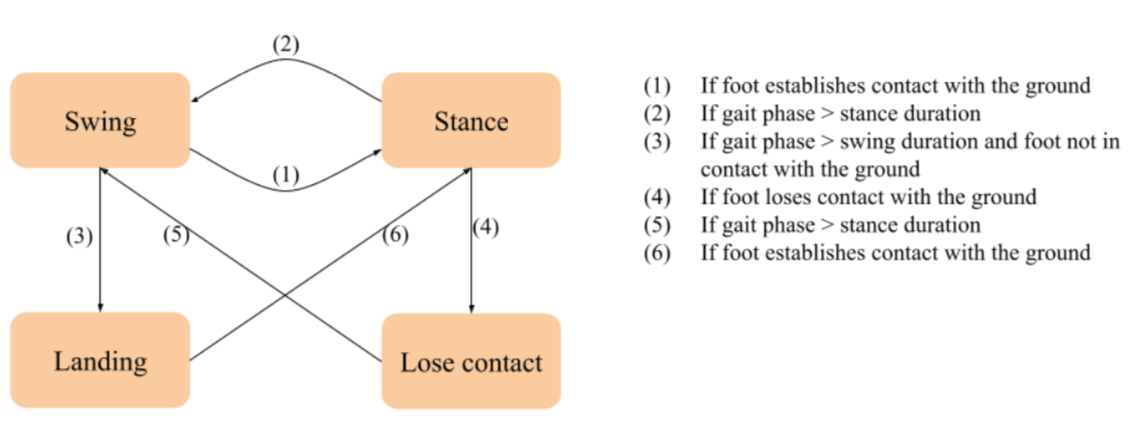}
    \caption{The FSM used in \protect\cite{yu_visual_2022} to control the state of each leg}
    \label{fig:laikago_fsm}
\end{figure}

In this work, the gait timings are fixed by a predefined gait scheduler, and cannot be altered by the learned policy. Different policies must be trained for different gaits. The gait scheduler sets the amount of time, and when, each leg is in swing and stance phase. By varying these times, different natural gaits such as trotting, walking, and galloping can be created. This gait scheduler makes use of the finite state machine (FSM) in Figure \ref{fig:laikago_fsm} that switches between swing and stance phases, in both a reactive and timing-based manner.

\subsection{Vision-based Policy}
\subsubsection{Observation and Action Space}
The vision policy is designed as a fully-connected neural network that  receives as input: depth images, CoM height, estimated CoM velocity and angular velocity, the robot’s current foot positions and contact states, the phase of each leg in the gait cycle, and the previous action as a form of smoothing. The policy then outputs the desired CoM pose and velocity for the next timestep, and the target landing $(x, y)$ location of each foot. An important note is that the $z$ component of the landing location is obtained directly from the depth image and is not learned. Interestingly, the authors find that leaving the policy to also learn the z location results in many invalid footstep locations and thus degrades its performance. 

The output of the neural network policy is also augmented in 2 ways. The nominal CoM velocity is actually calculated as a function of the chosen footstep location using an inverse of the Raibert SLIP heuristic, which relates foot position to desired CoM velocity. The output of the neural network is then added as a delta to this nominal velocity. 

Second, since 0 pitch is not necessarily always desirable (ie. in a scenario where the robot is climbing stairs or walking up a slope, zero pitch angle is actually suboptimal), the nominal pitch angle is calculated as a function of the difference between footstep heights of the front and rear stance feet heights. Thus, in a similar way to the CoM velocity, the pitch output of the learned policy is actually a delta from this nominal pitch angle. The authors report that these modifications are critical to good performance. 

\subsubsection{Reward Function}
An important component of successfully training an RL policy is the reward function. This work uses a relatively simple reward function:
\begin{align}
    R(s, a) &= clip(\dot{p_x}) - w_1(|p_y| + |\dot{p_y}|) - w_2|\Psi|
\end{align}

This function rewards high $x$ velocity (up until a set maximum, upon which it is clipped), penalizes $y$ velocity, and penalizes yaw angles. Having a low-level controller that already works enables this simple reward function to still train a performant policy. Later, we will see that methods that aim to learn more expressive policies require much more shaped rewards. 

The vision-based policy is trained using Augmented Random Search (ARS), which is relatively uncommon compared to other deep RL methods, but can quickly converge to a solution using MLP-based policies. The policy is trained for 2000 iterations of the ARS algorithm, which amounts to 1,024,000 total training episodes, with 256 perturbations of the policy being searched per training iteration. It is not reported in the work the wall-clock time required for training.

\subsubsection{Randomization}
Like other legged RL works, this method is trained using a physics simulation of the robot. A major hurdle for nearly all domains of robotic learning is that policies trained in simulation often do not perform well when deployed on real robots; this is often referred to as the sim2real gap. For policies that do not use visual input, a technique known as domain randomization is applied \cite{peng_sim_2018}, wherein certain parameters of the simulation and robot are randomly modified during training in order to ensure that the policy can succeed across a wide range of model variations. Noise is also often added to the inputs to the policy to add robustness. In this work, domain randomization is used to fine tune the policy after initial training to make it suitable to run on the real robot. The authors find that this results in faster training than applying domain randomization all throughout training.

\begin{figure}
    \centering
    \includegraphics[width=0.75\textwidth]{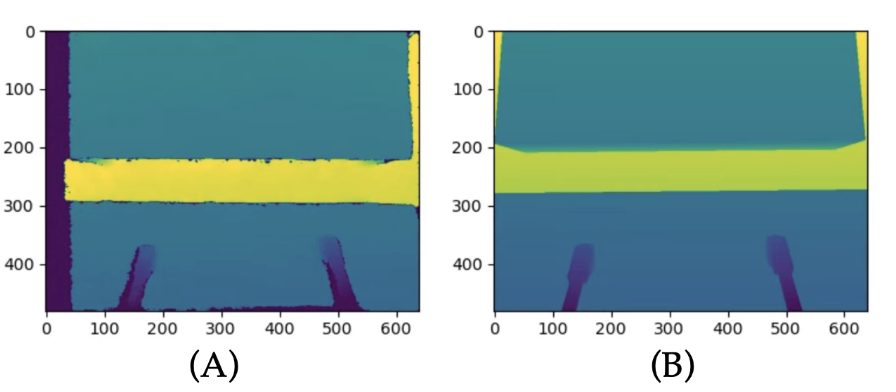}
    \caption{Real depth images (A) vs. simulated depth images (B)}
    \label{fig:visualloco_images}
\end{figure}

For visual inputs, domain randomization must be applied more thoughtfully. For example, simply adding Gaussian noise to depth images does not accurately mimic the noise seen in real depth images, as seen in Figure \ref{fig:visualloco_images}.

Despite the use of a learned visual policy, this paper also has a complex image pre-processing pipeline that is similar to the previous paper. To remedy the distribution shift between real and simulated depth images, the authors apply a multi-step filtering process to their simulated depth images.
\begin{enumerate}
\item First, Gaussian noise is added to the simulated image, and pixels are randomly blacked out.
\item Next, a Canny edge detector is applied to the simulated image to detect edges, and pixels close to edges are randomly dropped out. After these two operations, the simulated depth image more closely approximates the real images.
\item In both the real and simulated images, an inpainting operation is performed to fill in missing pixels.
\item Both the real and simulated are downsampled to reduce the effect of noise and missing pixels.
\end{enumerate}
Both this step and domain randomization are critical to the performance of the learned policy on the real robot. Unlike the parameter randomization, this image filtering pipeline is used all throughout training and at test time.

\subsection{Results}
\begin{figure}
    \centering
    \includegraphics[width=\textwidth]{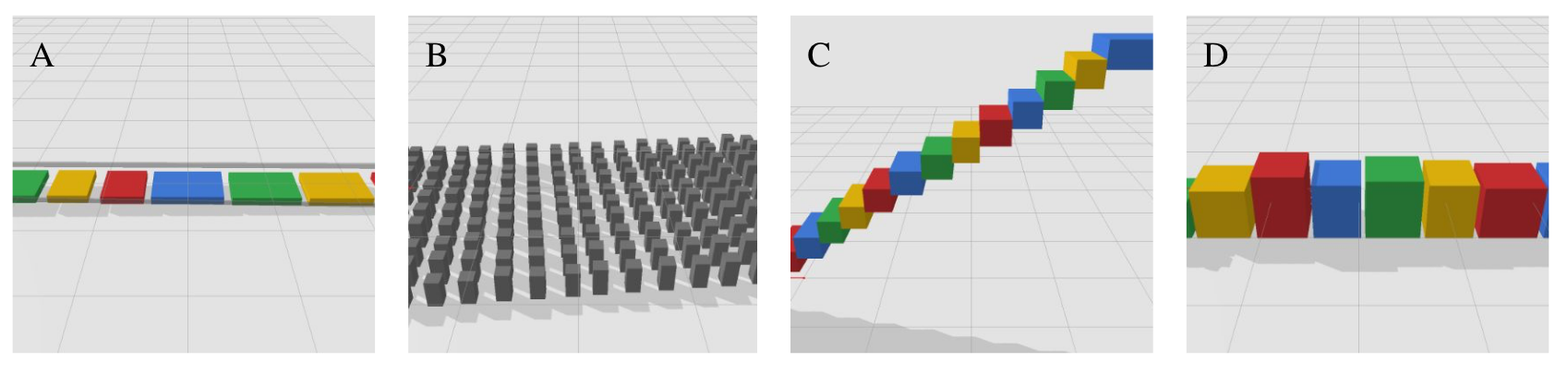}
    \caption{Some evaluation terrains from \protect\cite{yu_visual_2022}: from left to right, the terrain categories are stepping stones, quincuncial piles, stairs, and uneven terrain.}
    \label{fig:visualloco_terrains}
\end{figure}

The visual policy and motion controller are evaluated on various rough terrain scenarios in simulation. The types of terrains include stepping stones, randomized pillars, stairs, uneven terrains, and moving platforms. Some examples of these are shown in Figure \ref{fig:visualloco_terrains}. Due to random variations in terrains and perilous gaps scattered throughout, vision is necessary to successfully cross all terrains. It is worth noting that a separate policy is trained for each terrain type, and for each of 3 gait types: walking, trotting, and pacing. They report a performance ratio metric on each terrain, which is how far the robot traveled across the terrain compared to the maximum distance possible. Their method demonstrates strong ratios on all terrain types, traversing long distances on all. The quincuncial piles terrain (while walking) and moving platform terrains (while trotting) have the two lowest performance rates, but on average, the policy still makes it roughly 70\% across these terrains before failure.

\subsubsection{Evaluations in Simulation}
They also compare their visual policy method against some baselines commonly seen in the literature on the uneven terrain scenario. The first baseline is the heuristic foot placement method, similar to the previous paper. For this baseline, they train a CoM-only policy to output CoM positions and velocities, but not footstep locations. Instead, footstep locations are modified based on terrain unevenness criteria. Both the CoM targets and modified footstep are fed into the same motion controller as the full visual policy. The baseline with heuristic foot placements has worse performance, achieving about a 50\% performance ratio. A substantial reason for this is that the heuristic foot placement does not consider matching the footstep location with the planned CoM velocity, which makes the robot more unstable. 

The second baseline is a fully end-to-end policy, which maps depth images and robot states to joint angles. For this, they use the Policy Modulating Trajectory Generator architecture, which trains a neural network to modulate some nominal baseline trajectory. This architecture has been demonstrated to produce robust locomotion on the Minitaur and ANYmal legged platforms. However, in this case, the end-to-end approach does not achieve a high success rate on the uneven terrain scenario. The authors speculate that the end-to-end policy cannot learn precise foot placements since it only outputs target joint angles, and the uneven terrain scenario requires exact foot placement to avoid failure.

The last evaluation in simulation is a comparison to a blind policy that does not receive inputs from the depth cameras. Predictably, the blind policy achieves only a 10\% performance ratio, since the evaluation terrains have large features that require careful foot placement. While blind policies generally excel on more granular rough terrain, they cannot handle situations that require broader obstacle avoidance.

\subsubsection{Evaluations on a Real Robot}
Finally, the high level policy is evaluated on the real Laikago robot. The learned high-level policy, with some randomization fine-tuning, is able to successfully traverse the stepping stones scenario and climb up onto a pallet when deployed on a real robot. However, the policy does demonstrate some failures due to model mismatch between the simulated and real robots; it occasionally steps in the gaps between the stepping stones. Part of this is also due to the difficulty of achieving high-precision foot placement on a real moving robot. Because this is not an issue in simulation, the policy learns to step closer to edges than might be feasible on a real robot.

\subsection{Conclusions and Analysis}
From a practical standpoint, a major strength of this approach is the enhancement of an existing controller with ML in a relatively plug-and-play way. On all the perception-heavy tasks in this paper, the vision module presented a strict improvement over the blind policy. In cases that do not require perceptual input like walking on flat ground, the blind low-level controller still provides a robust baseline. Another advantage of this method is that it can deal with moving obstacles since it does not build a map, which could be advantageous when scaling up to environments with humans.

However, methods in this vein have a number of limitations that currently block real world deployment. The learning module must be retrained for both new terrain classes and new gaits. Other methods that use learning in locomotion have demonstrated strong generalization across different terrain types, so this modification is not fundamentally out of reach. However, some of the performance demonstrated in this work must be discounted since performance would be lost in trying to train a general policy.

Many works about learning locomotion appeal to the fact that hand-designing controllers requires a great deal of domain expertise, and that implementing a learned policy does not require as much tuning. Although this learned module is more expressive than the previous work’s vision module, it is still fundamentally limited by the performance of the low level controller. Therefore, the designers still have to spend time tuning the gains and formulating the MPC optimization to run well on real hardware. The low-level locomotion of the robot from the video demonstrations is also not particularly agile, although this is also in part due to the limitations of the low-cost Laikago robot hardware.

Overall this paper demonstrates a reasonably effective high-level policy learned on top of a low-level controller. Having a policy to modulate the speed of the robot in synergy with the selection of footsteps is certainly a strong idea and has the potential to be more expressive than \cite{kim_vision_2020}.

\section{Learning Robust Perceptive Locomotion in the Wild \protect\cite{miki_learning_2022}}
\label{sec:learningrobust}
The last work covered in this report \cite{miki_learning_2022} almost exclusively uses learned policies for locomotion. A learned blind locomotion controller forms the foundation of the policy to which perception is added to enable terrain adaptation. Fundamentally, this work has a similar aim to the previous two: it aims to improve the performance of the blind controller by integrating perception, while also accounting for unreliable inputs from the perception system. This work takes a “learn everything” approach to the problem, where neural networks are used for almost every significant computation and the amount of perception information used by the policy is also a learned function of the inputs. The policy is deployed on the ANYmal-C quadruped (Figure \ref{fig:anymal_terrains}) in a variety of natural environments. This work is exemplary of other end-to-end RL based methods in the literature, such as \cite{kumar_rma_2021}, and \cite{yang2022learning}.

\subsection{Training Scheme}
The policy is based on a teacher-student architecture for learning with privileged information. The teacher policy has access to noise-free sensor readings and extra world information from the simulator, and is trained using RL to output optimal actions with this full information. However, such a policy would be impossible to implement on a real robot since this extra information and noise-free perception is not available. So instead, the student policy is trained via supervised learning to mimic the actions of the teacher policy, but with incomplete and noisy information about the world at inference time. This kind of teacher-student training has been applied successfully to self-driving \cite{chen_learning_2019}.

Key components of the architecture that enable the student policy to learn from the teacher are the belief encoder and decoder modules, which estimate the true environment state from the incomplete and noisy data. These are essentially trained as denoising autoencoders, using reconstruction loss (squared L2 loss) between the true (privileged) state and the decoded latent vector. The output of the belief encoder, along with the commanded velocities and proprioception, is fed into an MLP which outputs the actions. During training, the teacher policy is run concurrently alongside the student policy, and the squared L2 distance between student policy’s output actions and the teacher’s actions is used to train the student policy.

\subsection{Observation and Action Spaces}
\subsubsection{Observations}
Both the teacher and student policies receive as input the commanded velocity (3-dimensional) and proprioceptive state of the robot. The proprioceptive state consists of the body orientation (3d) and velocity (6d), joint positions and velocities (12d each), a 3 time-step history of joint positions and velocities (36d each),  a 2-timestep history of the joint angle targets (24d), along with phase information from the CPG for each leg (13d total). All of these are concatenated together into a 133-dimensional vector.

The exteroception, is converted to a 2.5 heightmap constructed from the raw point cloud sensor data. The points in the heightmap are sampled in a spiral pattern around each foot. In total for all 4 feet, there are 208 sampled points. 

Only the teacher policy receives the privileged information, which consists of the contact states of each of the feet (a 4d binary vector), contact forces (a 12d vector), contact normals (a 12d vector), friction coefficients for each foot (4d), whether or not each thigh and shank are in contact with the terrain (8d vector), external forces and torques on the robot (6d vector), and “airtime”, which indicates how long each foot has been in the air (a 4d vector).

\subsubsection{Actions}
The action space for both policies is based on outputting deviations from a nominal Central Pattern Generator (CPG), which is a common paradigm in legged robotics \cite{ijspeert_central_2008}. The nominal CPG trajectory is a mapping from a periodic clock signal to a point in space that specifies where each foot should be at that point in the gait cycle. The clock signal is represented as an angle between 0 and 2pi, and the position of the foot is a vector in $\mathbb{R}^3$. Joint targets along this nominal trajectory are generated using inverse kinematics. The policy outputs a delta from the nominal phase and a delta from the nominal angle of each joint at that point in the gait cycle for each leg. These targets are then fed into a low-level PD controller that drives each joint to its target angle.

\subsection{Teacher Policy}
\begin{figure}
    \centering
    \includegraphics[width=0.75\textwidth]{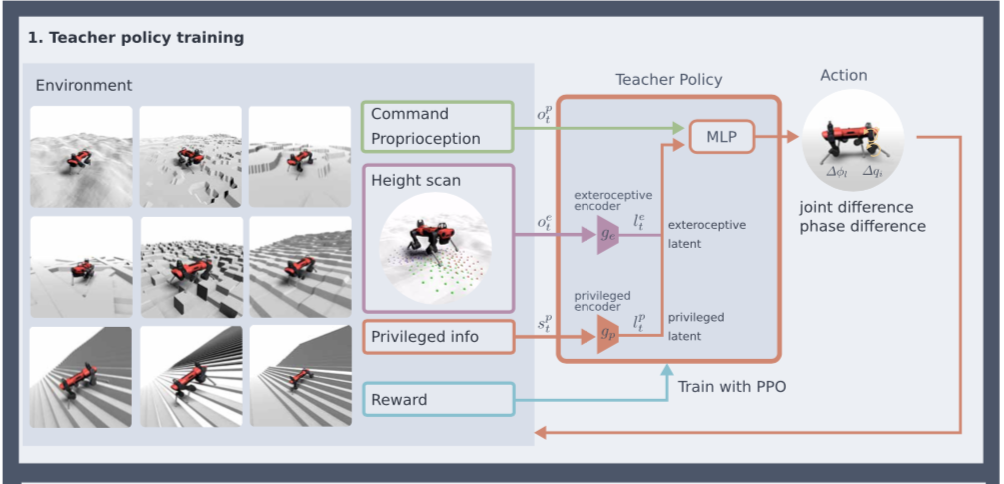}
    \caption{The teacher policy training setup from \protect\cite{miki_learning_2022}}
    \label{fig:anymal_teacher}
\end{figure}
\subsubsection{Reward Function}
The teacher policy is trained using RL with an extensively shaped reward to encourage it to learn stable motions. The reward function is a weight sum of 11 terms, each either encouraging desired behaviors or discouraging unwanted ones. These 11 terms are:
\begin{enumerate}
\item A linear velocity term, which gives maximum reward when the linear velocity of the body is close to the commanded velocity. The reward decays exponentially with respect to the distance between commanded and actual velocity.
\item An angular velocity term, which also gives maximum reward when the yaw velocity is close to the commanded yaw velocity. Again, this reward decays exponentially in the distance between the target and actual yaw velocity.
\item Orthogonal velocity reward, which gives a negative reward that scales in magnitude with any component of the velocity that is orthogonal to the commanded one.
\item Body motion reward, which penalizes components of motion not included in the command: any linear z, pitch, or roll velocity is penalized.
\item Foot clearance reward: this term encourages the swing foot to obtain as much clearance ($z$-distance) between the foot and the terrain below it as possible. However, this would cause the policy to always have unnaturally high swing foot motions, so this reward is instead formulated as a penalty if the swing foot is any less than 0.2m above the sampled terrain around it.
\item Shank and Knee collision reward: if any part of the leg other than the foot collides with the terrain, this term gives a negative reward that scales depending on how far along the policy is in training. The magnitude of this penalty increases as training proceeds.
\item Joint motion reward: This term penalizes joint velocity and acceleration in order to reduce unnecessary vibrations.
\item Joint constraint reward: this term penalizes the motion of the knee into a backwards configuration.
\item Target Smoothness Reward: This term penalizes high-magnitude first and second order derivatives of the foot motion, in order to encourage smooth motions.
\item Torque reward: This term penalizes high joint torques to minimize energy consumption.
\item Slip reward: This term penalizes any foot velocity (in the world frame) if the foot is in contact with the ground.
\end{enumerate}

\subsubsection{Randomization and Augmentation}
Besides the shaped reward, additional modifications to the training environments and terrains are needed to train a policy that will work on the real robot. Throughout training, the teacher policy sees many different types of terrain such as smoothly and discretely uneven, large steps, boxes, gridded steps, step stairs, and regular stairs. The difficulty of each of these is modulated by some noise parameters that control the magnitude and frequency of any deviations. In addition to the terrains, external forces and torques of variable magnitudes are applied to the robot during training, to simulate possible bumps and collisions the robot might encounter traversing the real world. Lastly, the masses of the robot’s body and legs are randomized during training to make the policy robust to model mismatch. Also noteworthy is that the authors train a model that captures actuator dynamics and latency on the real robot \cite{lee_learning_2020} and use this actuator model within the simulation.

\subsubsection{Training Procedure}
Applying fully difficult terrains and external disturbances to the robot from the start of training would hamper training by excessively penalizing exploration, so the authors adopt a curriculum structure for training the teacher policy. Curriculum training has been shown to improve the learning performance of legged locomotion policies by allowing greater initial exploration before making the environment too difficult \cite{iscen_learning_2021}. In this work, the curriculum modulates the difficulty of the terrain, the magnitude of external forces, and the penalty for collisions. The magnitudes of terrain and disturbances are kept small during the start of training, and are slowly ramped up as the policy becomes more capable, using a particle filter to estimate the policy’s capability and scale the difficulty appropriately. This curriculum factor also appears in collision and joint motion reward terms. Initially, the policy is allowed to have collisions and aggressive joint motions to encourage exploration, but as training continues the penalties are gradually increased. A diagram of the teacher training architecture and some of the example terrains seen in simulation is shown in Figure \ref{fig:anymal_teacher}.

The teacher policy is trained using Proximal Policy Optimization (PPO) \cite{schulman_proximal_2017}, an on-policy RL training algorithm. The teacher policy runs at 50Hz, which is a comparable action rate to other locomotion policies, and comparable to the high-level MPC and vision-based policy from the two previous papers. The robot is simulated in RaiSim, a closed-source rigid body physics simulator that implements a custom solver for fast contact dynamics. The fast solve time of RaiSim enables them to simulate 1000 agents in parallel, which speeds up data collection for PPO. The training time for the teacher policy is not reported in the paper. They report that this step is critical for sim2real transfer. Once the teacher policy has reached convergence, it is frozen and the student policy is trained. 

\subsection{Student Policy}
\begin{figure}
    \centering
    \includegraphics[width=0.75\textwidth]{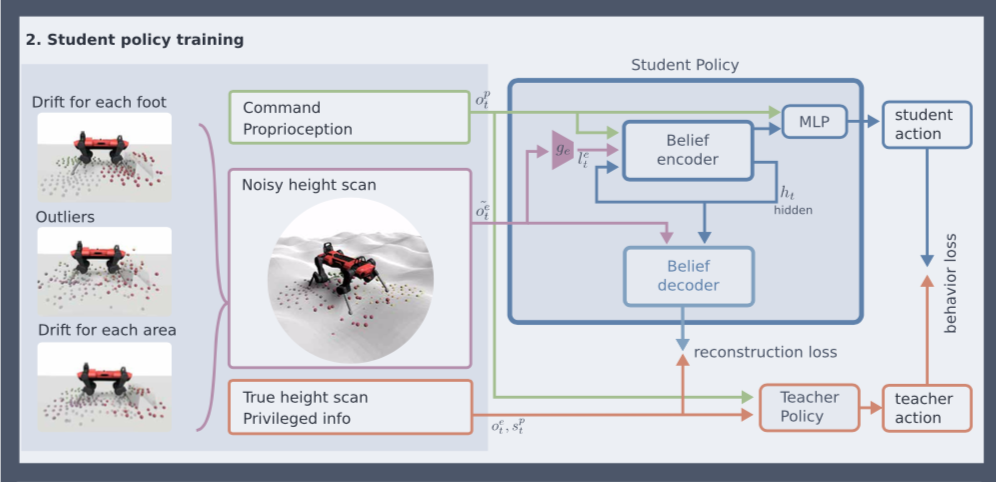}
    \caption{The student policy training setup from \protect\cite{miki_learning_2022}}
    \label{fig:anymal_student}
\end{figure}

The student policy receives the same command and proprioception data as the teacher policy, and also receives a noise-augmented pointcloud. The student policy does not receive any of the privileged data available to the teacher policy, since that data would not be available to a real robot. Because of the lack of privileged information and the noise added to the heightmap, the student policy faces a partially observed problem. This necessitates a different network architecture for the student policy. 

\subsubsection{Belief Encoder}
The key component of the student is a recurrent “belief encoder” and decoder. The belief encoder is a special type of recurrent network called a Gated Recurrent Unit (GRU). The belief encoder estimates a latent world state from a history of proprioceptive and exteroceptive observations. This conditions the learning problem for the student policy better since it is able to integrate past observations. In addition to the recurrent structure, the belief encoder is trained to not use exteroceptive information if it believes it is unreliable through a learned attention gate. The belief state at the current timestep is calculated from the proprioception and exteroception as follows:
\begin{align*}
    b_{\tilde{t}}, h_{t+1} &= \text{GRU}(o^p_t, l^e_t, h_t) \\
    \alpha &= \sigma(g_a(b_{\tilde{t}})) \\
    b_t &= g_b(b_{\tilde{t}}) + l_t^e \odot \alpha 
\end{align*}

Where $h_t$ is the previous hidden state of the GRU, $\alpha$ is the learned attention vector, $b_t$ is the final belief state, $o^p_t$ is the proprioceptive observation, and $l^e_t$ is the encoded exteroception. $g_a$, $g_b$, and $g_e$ are fully connected networks, and $\odot$ refers to the element-wise product. This belief encoder is trained using the standard autoencoder training method, in which the belief state is fed through a decoder that outputs estimated proprioception and exteroception vectors. These are then compared to the ground truth proprioception and exteroception using the squared L2 loss which is backpropagated through all the components of the decoder and encoder. The decoder is only used in the training process; only the computed belief state is used during inference and deployment.

The belief state and original proprioceptive input are then concatenated and fed into an MLP, which outputs the action. This action is compared against the teacher policy’s “optimal action”, and the squared L2 distance between the two is used to train the student policy. Thus, the student policy is trained mostly in a supervised and self-supervised way, and it learns to output optimal actions without having access to the same privileged information that the teacher policy has.

\subsubsection{Randomization and Augmentation}
The last major difference between the student and teacher policies is that the student policy receives a heightmap corrupted by shifting some of the scan points laterally, and perturbing the height values of some of the points. Both of these noise types are parameterized by a latent variable $z$ that encodes the variance of the noise. The values of $z$ generally represent three different regimes of operation:

\begin{enumerate}
\item Nominal noise that occurs when a good map is available during good operation
\item Large offsets that occur due to pose estimation drift or non-rigid terrain
\item Large noise magnitude to simulate sensor failure or occlusion
\end{enumerate}

These three noise regimes are sampled at a frequency of 60\%, 30\%, and 10\% respectively during training. This heightmap randomization helps to make the student policy robust to perception failures that occur during real world operation.

The student policy is also trained using a curriculum method. During the initial epochs of training, the student policy is fed noise-free heightmaps, and the variance of randomly sampled noise is gradually increased linearly as training progresses. In addition, the student policy only sees flat terrains for the first 10 epochs before other terrains are introduced through the same curriculum as the teacher policy. A diagram of the student policy training is shown in Figure \ref{fig:anymal_student}.

\subsection{Results}
\begin{figure}
    \centering
    \includegraphics[width=0.75\textwidth]{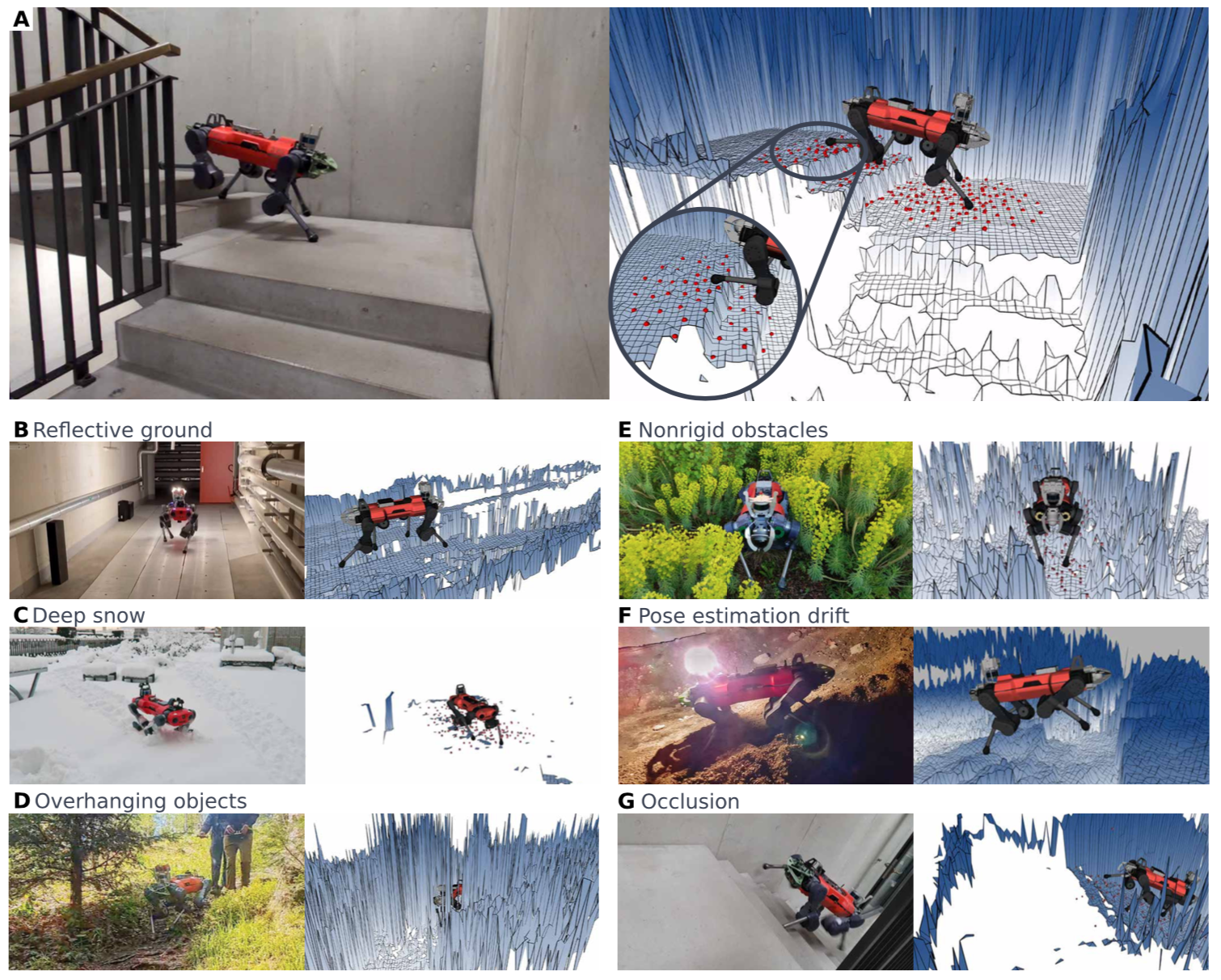}
    \caption{Some of the environments with degraded perception the ANYmal-C robot running the student policy was deployed in, and the reconstructed heightmaps.}
    \label{fig:anymal_terrains}
\end{figure}

The main focus of this work is the deployment of quadrupedal robots in unknown and rugged environments. To this end, the authors extensively evaluate the student policy in both indoors and outdoors scenarios. The student policy is trained and successfully deployed on the ANYmal-C robot in underground tunnels, forests with dense vegetation, snowy terrains, and indoor environments with stairs. In many of these terrains, the perception was degraded due to reflectance (from snow and glass), occlusions, overhanging objects, misleading inputs (for example, reporting soft vegetation as rigid terrain), and odometry drift, but the belief encoder and robust locomotion policy was able to overcome these challenges without falling. The authors report that their policy experienced zero falls while walking on these rugged terrains. Some examples of environments with difficult perception that the robot was deployed in are shown in Figure \ref{fig:anymal_terrains}

Perhaps the most novel evaluation compared to other locomotion work is a difficult 2.2 km hike to the summit of a mountain, which the ANYmal-C completed in 78 minutes. This hike involved walking on inclinations over 38\%, crossing rocky and wet surfaces, and stepping over tree roots and vegetation. Despite the challenging locomotion scenario, the robot is able to complete the hike without falling over. In all of these evaluations, a human operator is providing joystick commands, so the robot is only operating semi-autonomously.

The next evaluation is the contribution of exteroception to the overall locomotion performance. To measure this, the perceptive policy is compared in many terrain scenarios against the authors’ previous work \cite{lee_learning_2020} that is a blind locomotion policy . The perceptive policy is able to climb stairs up to a height of 32cm, at which point it stops because it instead starts to treat the steps as obstacles. Meanwhile, the blind baseline can only reliably ($\geq50\%$ success rate) climb steps that are 17cm tall. One failure case for the perceptive policy in this case was that it tried to go around steps rather than climb up them. The next evaluation is an obstacle course, which each policy tries to follow using a pure pursuit controller. The obstacle course consists of steps, inclined platforms, a raised platform, and piled blocks. The perceptive policy allowed the robot to adapt and plan for the obstacles, resulting in smooth tracking throughout the obstacle course. Meanwhile, the baseline failed to track the desired path.

The integration of perception into the policy also allows it to move faster. During training, the blind locomotion policy is exposed to environments with rough terrain that it cannot perceive. Meanwhile, the perceptive policy is able to verify that the ground is actually flat and therefore achieves faster locomotion speed than the blind policy even on flat terrain. Although they compared to the proprioceptive baseline in some indoors environments, it would have further solidified their thesis if they also compared it against the proprioceptive baseline in the same outdoors environments.

Lastly, the authors evaluate the robustness of the policy to misleading perception. The authors evaluate the policy in 4 scenarios where perception is misleading: walking over a soft foam block, walking on a glass block, covering the perception sensors with tape, and manually inducing kinematic estimation error so the point cloud is sampled in the wrong location. In all of these examples, they decode the belief state to visualize how it is updated in the presence of these uncertainties. The figure below shows how, for the foam block, the decoded belief state first estimates the terrain to be at the height of the foam, but after the robot steps through it, the decoded belief state reflects that solid ground is actually below the foam. More importantly, due to the recurrent structure, this belief state consistently retains this knowledge in the future. A similar example is shown for the glass block and for a slippery step, where the estimated friction coefficient drops after the robot slips on the stairs.

\subsection{Conclusion and Analysis}
Due to the relatively unconstrained action space (effectively, the policy can output joint angle targets for every joint on the robot), the policy trained in this work has the capability to learn the most expressive controller out of the three methods. However, this freedom also comes with some costs. RL methods notoriously get stuck in local minima while training, especially when using perception as input. This necessitates extensively shaping the reward and inputs the policy sees while training. The reward function contains 11 terms which reward desirable components of motion and penalize undesirable ones. Furthermore, the nominal (zero) action is based on a CPG, which will significantly bias the training and prevent the policy from learning more agile motions. Another downside that comes along with a fully learned policy trained with RL is the extensive tuning of hyperparameters, reward weights, curriculum terms, and noise distribution terms that is required to obtain good performance, and there is no way to alter the behavior of the policy without fully retraining it.

This method does have many upsides though. The use of the belief encoder is an interesting and powerful way of denoising perception and estimating extrinsics that are not directly observable, similar to \cite{kumar_rma_2021}. The heightmap is created as an intermediate representation from the perceptive input, which could be from a lidar or a set of depth cameras. Processing the constructed 2.5D heightmap using neural networks can render the method more robust to lighting variations that the filtering methods of \cite{kim_vision_2020} and \cite{yu_visual_2022} might fail on. This robustness is demonstrated through the extensive outdoor evaluations. However, this comes at the cost of requiring extensive curriculum design and noise augmentation during training. The randomization also induces some bias into the denoising autoencoder: one problem in particular is that occlusions such as deep holes around stepping stones cause the encoder to assume that there is a sensor failure, and the belief encoder will fill in those holes as flat terrain. This is a general problem with learning systems; opaque biases that appear during training can cause the final policy to behave in unpredictable ways. Especially with RL, these biases can be difficult to root out.

Overall, despite some flaws, this is a powerful method for perceptive locomotion. The extensive training curriculum and the teacher student architecture enables robust locomotion in a broad variety of environments. Furthermore, there is a clear demonstration that perception strictly improves the robot's performance.

\section{Comparisons}
\label{sec:comparison}
\subsection{Performance and Autonomy}
Ultimately, the goal of incorporating perception into the control stack of legged robots is to enable greater autonomy by eliminating collisions and falls. In this section, I will analyze how each method approaches that goal by examining their abilities to overcome rough terrain and the level of human supervision each method requires during deployment. 
	
The three-stage pipeline shown in \cite{kim_vision_2020} shows some rudimentary obstacle avoidance by using potential fields and it is able to modify its footstep locations to avoid walking on edges and impassable regions. However, classifying each of these requires tuning some heuristic factors to decide which terrain slopes are unstoppable or impassable, which could make the method undesirably conservative or aggressive. The evaluations in the paper show some ability to cross some rigidly uneven terrain, but were only conducted indoors due to the tether, so the ability to deal with meaningful obstacles outdoors is unclear. Another limiting factor in this is that the parameters of the swing foot trajectory are not modified using perception. Although it queries the heightmap for the z location of the footstep, it could still trip over small hazards on the ground like branches or stones since the maximum height of the swing foot is fixed and it cannot alter the fixed contact schedule. Besides that limitation, the underlying low-level controller shows the ability to trot at 1 m/s across gravel and wet grass and 3.7 m/s on flat ground indoors, so there is some promise for robustness in these different scenarios. Alongside this, their method allows a somewhat neat integration of motion libraries, which allow the designers to leverage highly dynamic pre-optimized motions that do not necessarily fit a normal locomotion paradigm into the control stack. This shows strong potential in terms of what a well-integrated perception module could achieve with this controller. If the simple depth map filtering scheme and SLAM can properly work in outdoors environments, this method should be able to handle some limited rough terrains.
	
The learned high-level policy from \cite{yu_visual_2022} fixes some of the problems in the previous three-stage hierarchical pipeline. Because it learns footstep locations through trial and error on the robot, it does not require setting heuristic thresholds on terrain gradients. Such a method can achieve precise foot placements by using knowledge of the robot model, allowing it to traverse the quincuncial piles terrain (in simulation). In addition, the method can learn to synergize its footstep locations with the planned center of mass velocity, resulting in more stable motions on rough terrain. Additionally, since it does not require mapping and localization, it can also react to moving obstacles like humans, cars, or other robots. However, it still suffers from the inability to adjust its swing foot parameters and the gait schedule. Furthermore, their method does not show generalization to different terrain scenarios than those on which it was trained, which greatly limits the potential to handle many varied terrains. However, other similar works do show the ability to handle different terrain types, so this is something that could be resolved through changes in the learning method like introducing a curriculum or using domain randomization throughout training.
	
The fully-learned policy on ANYmal-C demonstrates stable locomotion on the widest range of terrains out of the three works, including underground tunnels, forest trails, snowy steps, and more. Because it has the most freedom to learn different motions, it can adapt to a wide variety of obstacle types, and the extensive perception augmentation and domain randomization makes it robust enough to handle slippage and unseen obstacles. This is a key advantage of having such an expressive action space. The fully learned policy is not restricted to any particular gaits, whereas the MiniCheetah and Laikago works only work with predefined gaits. This gives the ANYmal-C method the freedom to adapt its contact sequence to the terrain. In addition, the recurrent “denoising” scheme also helps the robot walk through areas where perception might be misleading, such as tall grass. Either of the two previous methods would not be able to handle such scenarios with their relatively simple filtering schemes and limited training environments. However, this fully learned policy also has its own limitations. Because it only learns joint angle policies, it may not be able to handle tasks where precise foot placement is required (such as stepping stones or narrow stairs), and they do not perform any evaluations in these kinds of scenarios.

One noted phenomenon in legged robots is that an expert human operator is able to guide the robot through difficult scenarios even with just high level twist commands \cite{gibson_terrain_2022}. All three of these methods work towards approaching the level of the human expert pilot. Although the rough terrain capabilities differ between the three methods, they actually offer a similar level of autonomy. All three require a human operator to provide a velocity or heading command. The MPC+WBIC framework in \cite{kim_vision_2020} is designed to follow a planned trajectory, while the ANYmal learned framework learns to follow a linear and yaw velocity. The high-level policy on Laikago can only walk in a straight line by default (it is only trained to do so); in order to turn, a yaw angle command must be given to the MPC controller by a human operator. All of these methods enable low-level autonomy in that the planner perhaps does not have to react to small obstacles or adjust its commands to the terrain, but they still require a high-level planner to provide velocity commands. Designing a planner with low-level model knowledge that can utilize the agility of legged robots in the same way that human operators can is still a significant challenge, although some strides have been made in other works \cite{norby_fast_2020,fu_coupling_2021}.

\subsection{Information Bottleneck}
One of the main philosophical differences we observe between these three methods is the bottleneck in how much information flows from perception to action. \cite{kim_vision_2020} use perception on the MiniCheetah in a very limited capacity: simple features are extracted from the heightmaps for obstacle avoidance and guiding footstep planning. \cite{yu_visual_2022} employ a wider bottleneck both in terms of features and action space. It uses learned features from perception to also inform the Laikago CoM trajectory alongside the footstep selection. \cite{miki_learning_2022} allow the most information to flow since the perception to action mapping is totally unrestricted aside from a learned modulation factor. 

Each of these methods has its own advantages and disadvantages. Both the MiniCheetah and Laikago work allow the perception module to slot in neatly on top of an existing, high-performing blind controller, and do not require any modifications to the controller itself. If the low level controller is shown to be robust, it is unlikely that perceptive noise and failures will greatly reduce the performance or the stability of the policy in scenarios where perception is not critical. This separation would also allow independent validation of each of the perception and control modules to ensure their safety. This comes with the tradeoff that the performance enhancement from perception is still fundamentally limited by design of the low-level controller.

Meanwhile, in the method on ANYmal-C, there is no distinct “perception module” that exclusively processes perceptive inputs. Instead, perception and action are tightly coupled in the learned framework. This comes with the converse tradeoffs to the two previously mentioned works. The lack of bottleneck between perception and action enables all available perception information to be utilized by the controller, and in turn enables faster locomotion even on flat ground since the controller is aware that there are no nearby obstacles. This also holds in reverse; if perception is truly unreliable, it can greatly reduce the performance of the controller, and an explicit safety mechanism in the form of the belief encoder has to be designed to prevent this failure mode.

\subsection{Perception Uncertainty}
Although the performance of all three methods is enhanced by perception, depth cameras and lidars are not always reliable; images and point clouds are corrupted by noise, position drift, and occlusions. All vision-based pipelines must deal with these challenges. 

Both \cite{kim_vision_2020} and \cite{yu_visual_2022} make use of a dilation/inpainting filtering scheme to deal with outlier points and holes in their point cloud scans. The second method also makes use of a low-pass filtering operation (downsampling the image) to deal with noise. Both of these schemes work well in their indoor environments with relatively rigid, planar obstacles. The challenge of these hand-engineered perception systems has always been how they handle situations that they were not designed for, and this also applies to both of these works. It remains to be seen how well both of these filtering systems would work in more unstructured environments, dealing with fine-grained obstacles such as tree roots and lighting variations caused by sunlight. Another downside of \cite{kim_vision_2020}, with respect to perception uncertainty, is the use of a localization and mapping pipeline to plan footsteps in the global frame. Simultaneous Localization and Mapping (SLAM) systems can be unreliable due to integration drift and keypoint matching errors, especially when used on legged robots due to the constant impacts and shaking of the body. Although they demonstrate their low-level controller without perception in outdoors terrain, it remains yet to be seen how this could scale to outdoors experiments. 

Both \cite{yu_visual_2022} and \cite{miki_learning_2022} make use of learning and augmentation to deal with these problems. Learning methods have shown extremely strong results in various computer vision tasks, such as object recognition, denoising, and segmentation. Therefore, using learning to process the depth scans shows promise in dealing with noise in perception. This might allow the perception scheme in \cite{yu_visual_2022} to overcome some of the issues discussed in the previous paragraph, and enables good performance in \cite{miki_learning_2022}. However, the belief encoding compression scheme in the latter also does have some downsides. It cannot handle occlusions well; the belief encoder scheme assumes that because there are no points sampled in that area, that the terrain is continuously flat in that area. This can cause catastrophic failures in the presence of deep gaps or cliffs. Furthermore, because the point cloud is registered around each foot, it relies on good kinematic pose estimation for each foot. If there is pose estimation error, the point cloud will be incorrectly sampled and cause errors. 

\subsection{System Complexity and Engineering Effort}
A major consideration for system design is how much expertise, hand-tuning, and design effort are required to successfully implement these methods on real systems. Although this aspect is not usually reported in legged robotics works, it is a big determinant of whether or not a method will be used in practice. One of the main arguments for using machine learning in legged locomotion is that less domain expertise is required to obtain a functioning controller. Generally, methods that use learning do require less expertise with legged systems, with some major caveats.

The work on the MiniCheetah is an example of a typical hierarchical pipeline where each component requires significant manual expertise with legged robots. The MPC and WBIC controllers require extensive knowledge of how to formulate optimization problems and the dynamics of legged locomotion. In addition, to actually work on the robot the pipeline required the efficient computation of the local inertia matrix $A$ and Coriolis force $b$, a Kalman-filter for estimation of body position and velocity, a manually-specified gait scheduler, and the offline design and optimization of the jumping motion. 

Besides the SLAM system, the high-level perception module of \cite{kim_vision_2020} is also relatively simple, and does not require much hyperparameter tuning aside from setting thresholds for the different terrain classifications. Given the simple nature of the vision system and the off-the-shelf SLAM system, if the designer already has a low-level controller implemented, this approach is an appealing method to implement perceptive locomotion without extra modifications to the controller. However, as pointed out earlier in this report, the SLAM system can be a major source of error, especially for a small robot on the scale of the MiniCheetah, and might require significant tuning to work in different environments and lighting conditions.

In the ANYmal-C work, we saw an extensively shaped reward function with different 11 terms with tuned weights (for example, the torque, smoothness, and slip penalties). All terms besides the velocity command reward require some knowledge about what traits are desirable in legged locomotion. In addition, the method requires designing a curriculum and randomization scheme, along with hyperparameter tuning for each of these. When deploying this method on different robot geometries, it is unclear how much retuning and redesigning would be required. A few of the reward terms used in the ANYmal-C work are also specific to the ANYmal-C itself, meaning that if deploying this method on a different robot, some retooling and retuning would be required. Another aspect that the authors claim is critical for sim2real generalization of their policy is learning an actions-to-torques model for how the actuators behave and using that model in the simulation of the robot. This step requires designing a data collection scheme from the real robot.

Although the ANYmal-C learned policy does not require a hand-designed low level controller, this also means that there is no way to alter the locomotion behavior of the robot without retraining the entire policy or incorporate more agile motions. Meanwhile, the behavior of the MiniCheetah and (to a lesser extent) the Laikago can be altered by changing task specifications, gains in the optimization problems, and the gait schedules.

The work on Laikago straddles a line between \cite{kim_vision_2020} and \cite{miki_learning_2022}. Implementing a high-level policy requires manually designing a low-level controller and specifying the action space for the high level policy. This additional effort comes with the benefit of not having an extensively shaped reward function: the reward function contains only a few terms, all of which are directly tied to the objective of the robot to walk forward. However, even with the simplified action space, the Raibert heuristic augmentation is critical to the performance of the policy. Since this work also uses RL, they also have to employ randomization and augmentation to handle the sim2real gap, along with some tuning of the terms of the reward function and hyperparameters of the policy. 

Another practicality factor with learning-based policies is the iteration time between experiments. Both the Laikago work and ANYmal-C work require significant training time, on the order of millions of training episodes. Both use on-policy RL algorithms, which are generally known to be less sample efficient than their off-policy counterparts. Since simulators have grown increasingly capable and a number of works have shown “zero-shot” sim2real transfer for legged robots, sample efficiency has become less of a concern. However, the hours spent performing validation and hyperparameter tuning add a significant delay to development time. Based on all of these factors, choosing an RL solution for locomotion does not automatically guarantee lower development time.

\section{Future Work}
\label{sec:futurework}
Although each of these methods makes strides in terms of advancing perceptive locomotion, there is still extensive future work, both in each of the veins of the individual works, and at a thematic level. 

In the vein of the MiniCheetah work and other heavily model-based methods, further work needs to be done so the perceptive locomotion can approach the agility of the blind controller on flat ground. This could involve extending the action space to allow perceptive input to modulate other parameters of the locomotion or incorporating some learned encoder to extract more of the information available through perception. The use of the jumping motion library into the method provided an interesting way of crossing otherwise impassable terrains; adding more jumps or motion types into the library would also greatly extend the capabilities of this method. In general, more experimentation on other legged platforms and terrain types would also validate the robustness.

For \cite{yu_visual_2022}, immediate future work would be to train a single policy that can work on multiple terrains types and gaits. In addition, a more robust lower level controller, such as the WBIC from the previous work, could be integrated into the control framework. A few avenues to improve the expressiveness of the policy could also be pursued. Incorporating terrain textures (ie. not just depth) could also improve the vision based policy in outdoors environments. Lastly, the action space of the vision policy could also be extended to also actively change the gait when it detects that one type would be more optimal for a particular terrain.

Although \cite{miki_learning_2022} demonstrate the best performance on natural terrains out of the three, there are still significant extensions possible. In terms of the locomotion, the method could be extended to more agile modes than the CPG-based locomotion, such as jumping or galloping. On the perceptual side, incorporating more environment information, such as texture or lookahead, would give the learner more information and allow it to adapt to even more environment types. Furthermore, explicitly using the uncertainty in the belief state could give the policy more information about when to be aggressive versus when to be tentative and could prevent the problem of the encoder making a flat terrain assumption for occluded areas. Lastly, evaluation in environments with moving obstacles would validate that this method could work in human environments.

Across all three methods, some general trends emerge for future directions in research. Robustness has always been a driving concern in legged locomotion research. Ensuring this robustness, especially in learning-based methods, generally implies less agility. For example, although the MiniCheetah blind controller is able to achieve speeds of up to 3.7 m/s on flat ground, the stack with perception is only run at 0.38 m/s. The ANYmal-C perceptive policy also only reaches a max speed of 1.2 m/s. One application of perception can be to enable high-level adaptation to different environments and fine-tuning for greater agility depending on the scene it observes through perception. This adaptation could occur through changing motion parameters, optimization gains, or commanded velocities for the non-learned controllers, or a meta-learning framework for the RL-based methods. Incorporating semantic information into the locomotion policies could also be another avenue for achieving this.

In addition, long-horizon planning to reach goals or waypoints is critical for true autonomy. All three of these works generally aim to make the controller as capable as possible to simplify the planning problem. In more challenging environments, the planner must be able to understand the capabilities of the robot and controller. Although some strides have been made in this area, developing a long-horizon planner that can capitalize on the dynamic capabilities of legged robots is still an open challenge.

\section{Conclusion}
\label{sec:conclusion}
Upon comparing each of the three methods, a few broad tradeoffs emerge between the various options. Throughout legged robotics, there is a general tradeoff between the interpretability and robustness of model based methods and the performance capabilities of learned methods. From a system design perspective, more structured perception systems are easier to design and integrate with an existing control stack, but they make more assumptions and are limited in their expressiveness, performance, and generalization. For example, the perception system on MiniCheetah was relatively simple to implement, but was restricted in that it could not alter the magnitude of the commanded velocity or contact schedule and only used gradient information from the terrain. From a robustness perspective, learned perception systems are better able to handle nominal levels of noise, but suffer from the long tail problem and behave unpredictably when faced with novel situations. Model-based  or control theoretic methods give better guarantees and behave more predictably, but are capped in terms of the performance and agility they can achieve. On the other hand, model-free methods require extensive randomization and curriculum engineering to achieve realistic motions, and their behavior cannot be easily altered online or be used on other robot geometries. How to select between each of these tradeoffs is highly dependent on the specific use case of the designer. Each of the methods presented in this report provides a viable solution for integrating perception into locomotion, and each brings the promise of general-purpose, field-deployed legged robots closer to fruition.

\pagebreak

\bibliographystyle{apalike}
\bibliography{primary}
 
\end{document}